\documentclass{llncs}

\usepackage{graphicx}
\graphicspath{{./figs/},{./plots/}}
\usepackage{caption,subfig}
\usepackage{epstopdf}
\usepackage{amsmath}
\usepackage{mathptmx} 
\usepackage{cite} 

\begin{document}

\mainmatter              

\title{DNA Reservoir Computing:\\ A Novel Molecular Computing Approach}

\titlerunning{A Novel Molecular Computing Approach: DNA Reservoir Computing}

\author{Alireza Goudarzi\inst{1}, Matthew R. Lakin\inst{1}, \and Darko Stefanovic\inst{1,2}}

\authorrunning{Alireza Goudarzi et al.} 

\institute{Department of Computer Science\\ University of New Mexico\\ \and Center for Biomedical Engineering\\ University of New Mexico\\ \email{alirezag@cs.unm.edu}}

\maketitle         

\captionsetup{font=small}

\begin{abstract}
We propose a novel molecular computing approach based on reservoir computing. In reservoir
 computing, a dynamical core, called a \protect{\em reservoir},  is perturbed with an external input
  signal while a \protect{\em readout layer} maps the reservoir dynamics to a target output.
Computation takes place as a transformation from the input space to a 
high-dimensional spatiotemporal feature space created by the transient 
dynamics of the reservoir. The readout layer then combines these features to produce the target output.
 We show that coupled deoxyribozyme oscillators can act as the reservoir. We show that despite
  using only three coupled oscillators, a molecular reservoir computer could achieve $90\%$
   accuracy on a  benchmark temporal problem.


\end{abstract}

\section{Introduction}
A reservoir computer is a device that uses transient dynamics of a system in a critical regime---a
 regime in which perturbations to the system's trajectory in its phase space neither spread nor die out---to transform an input signal into a desired output~\protect\cite{Jaeger02042004}. We propose a novel 
 technique for molecular computing based on the dynamics of molecular reactions in a microfluidic setting.
  The dynamical core of the system that contains the molecular reaction is called a reservoir. We design  
  a simple {\em in-silico} reservoir computer using a network of deoxyribozyme 
  oscillators~\protect\cite{Farfel2006}, and use it to solve temporal tasks. The advantage of this
   method is that it does not require any specific structure for the  reservoir implementation except
    for rich dynamics. This makes the method an attractive approach to be used with emerging
    computing architectures~\protect\cite{springerlink:10.1007}.

We choose deoxyribozyme oscillators due to the simplicity of the corresponding mathematical model
 and the rich dynamics that it produces. In principle, the design is generalizable to any set of  reactions
  that show rich dynamics. We reduce the oscillator model in~\protect\cite{Farfel2006} 
to a form more amenable to mathematical analysis. Using the reduced model, we show that the oscillator
 dynamics can be easily tuned to our needs. The model describes the 
oscillatory dynamics of three product and three substrate species in a network of three coupled oscillators.
 We introduce the input to the oscillator network by fluctuating the supply of 
substrate molecules and we train a {\em readout layer} to map the oscillator dynamics onto a target output. For a complete physical reservoir computing design, two main problems should be addressed: (1) physical implementation of the reservoir and (2) physical implementation of the readout layer. In this paper, we focus on a chemical design for the reservoir and assume that the oscillator dynamics can be read using fluorescent probes and processed using software. We aim to design a complete chemical implementation of the reservoir and the readout layer in a future work (cf. Section~\protect\ref{sec:discuss}). A similar path was taken by Smerieri et al. \protect\cite{NIPS2012_0456} to achieve an all-analog reservoir computing design using an optical reservoir introduced by Paquot et al. \protect\cite{Paquot:2012uq}.

We use the molecular reservoir computer to solve two temporal tasks
of different levels of difficulty. For both  tasks, the readout layer must compute a function of past inputs
  to the reservoir. For Task A, the output is a function of two immediate past inputs,
 and for Task B, the output is a function of two past inputs, one $\tau$ seconds ago and the other
  $\frac{3}{2}\tau$ seconds ago. We implement two varieties of reservoir computer, one in 
which the readout layer only reads the dynamics of product concentrations and another in which
 both product and substrate concentrations are read. We show that the product-only version 
achieves about $70\%$ accuracy on Task A and about $80\%$ accuracy on Task B, whereas the
 product-and-substrate version achieves about $80\%$ accuracy on Task A and 
$90\%$ accuracy on Task B. The higher performance on Task B is due to the longer time
 delay, which gives the reservoir enough time to process the input. Compared with other reservoir
  computer implementations, the molecular reservoir computer performance is surprisingly
 good despite the reservoir being made of only three coupled oscillators.

\section{Reservoir Computing}
\label{sec:rc}

 As reservoir computing (RC) is a relatively new paradigm, we try to convey the 
sense of how it computes and explain why it is suitable for 
molecular computing. RC achieves computation using the dynamics of an excitable 
medium, the reservoir \protect\cite{Lukosevicius:2009p1443}.  
We perturb the intrinsic dynamics of the reservoir using a time-varying input and 
then read and translate the traces of the perturbation on the system's 
trajectory onto a target output. 

RC was  developed independently by Maass et al. 
\protect\cite{Maass:2002p1444} as a model of information processing in cortical 
microcircuits, and by Jaeger \protect\cite{Jaeger:2002p1445} as an alternative 
approach to time-series analysis using Recurrent Neural Networks (RNN). In the 
RNN architecture, the nodes are fully interconnected and 
learning is achieved by updating all the connection weights 
\protect\cite{Widrow:1990p2053,Jaeger:2002p1445}. However, this process is 
computationally very intensive. Unlike the regular structure in RNN, the 
reservoir in RC is built using sparsely interconnected nodes, initialized with fixed random 
weights. There are input 
and output layers which feed the network with inputs and obtain the output, respectively. 
To get the desired output, we have to compute only the weights on the 
connections from the reservoir to the output layer using examples of input-output sequence.

 \begin{figure}[t]
\centering
\includegraphics[]{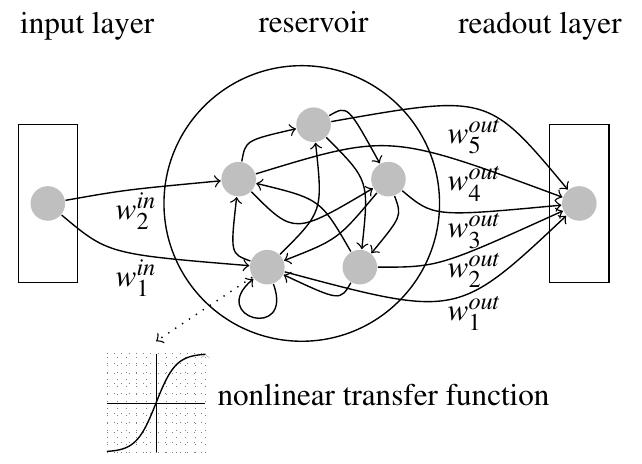}
\caption{Schematic of a generic reservoir computer. The input is weighted and 
then fed into a reservoir made up of a number of  nodes with nonlinear 
transfer functions. The nodes are interconnected using the coupling matrix 
${\bf W}^{res}=[w^{res}_{ij}]$, where $w^{res}_{ij}$ is the weight from node $j$ to 
node $i$. The weights are selected randomly from identical and independent 
distributions. The output is generated using linear combination of the 
values of the nodes in the reservoir using output weight vector  
${\bf W}^{out}=[w^{out}_i]$. }
\label{fig:fig_1}
\end{figure}

Figure~\ref{fig:fig_1} shows a sample RC architecture 
with sparse connectivity between the input and the reservoir, and between the nodes 
inside the reservoir. The output node is connected to all the reservoir nodes. 
The input weight matrix is an $I
\times N$ matrix ${\bf W}^{in}=[w^{in}_{i,j}]$, where $I$ is the number of input 
nodes, $N$ is the number of nodes in the reservoir, and $w^{in}_{j,i}$ is the 
weight of the connection from input node $i$ to reservoir node $j$. 
The connection weights inside the reservoir are represented by an
 $N\times N$ matrix ${\bf W}^{res}=[w^{res}_{j,k}]$, where $w^{res}_{j,k}$ is the 
weight from node $k$ to node $j$ in the reservoir. The output weight matrix is 
an $N\times O$ matrix ${\bf W}^{out}=[w^{out}_{l,k}]$, where $O$ is the number 
of output nodes and $w^{out}_{l,k}$ is the weight of the connection from 
reservoir node $k$ to output node $l$. All the weights are samples of 
i.i.d. random variables, usually taken to be normally distributed with mean $\mu=0$ and 
standard deviation $\sigma$. We can tune $\mu$ and $\sigma$ depending 
on the properties of $U(t)$ to achieve optimal performance. We represent the 
time-varying input signal by an $I$th order column vector ${\bf U}(t)=[u_i(t)]$, the 
reservoir state by an $N$th order column vector ${\bf X}(t)=[x_j(t)]$, and the 
generated output by an $O$th order column vector ${\bf Y}(t)=[y_l(t)]$. We  
compute the  time evolution of each reservoir node in discrete time 
as:

\begin{equation}
x_j(t+1)=f({\bf W}^{res}_j\cdot {\bf X}(t)+{\bf W}^{in}\cdot {\bf U}(t)),
\end{equation}
where $f$ is the nonlinear transfer function of the reservoir nodes, $\cdot$ is the matrix dot product,
 and ${\bf W}^{res}_j$ is the $j$th 
row of the reservoir weight matrix. The reservoir output is then given by:
\begin{equation}
{\bf Y}(t)=w_b+{\bf W}^{out}\cdot{\bf X}(t),
\end{equation}
where $w_b$ is an inductive bias. One can use any regression method to train the output weights to 
minimize the output error $E=||{\bf Y}(t)-{\bf \widehat{Y}}(t)||^2$ given the target output 
${\bf \widehat{Y}}(t)$. We use linear  regression and calculate the weights using the  Moore-Penrose pseudo-inverse method \protect\cite{PSP:2043984}:
\protect\begin{equation}
{\bf W}^{out'} =({\bf X'}^T\cdot{\bf X}')^{-1}\cdot{\bf X}'^T\cdot {\bf Y'}.
\end{equation}
Here, ${\bf W}^{out'}$ is the output weight vector extended with a the bias
 $w_b$, {\bf X'} is the matrix of observation from the reservoir state where each
  row is represent the state of the reservoir at the corresponding time $t$ and
   the columns represent the state of different nodes extended so that the last column
    is constant 1. Finally, ${\bf \widehat{Y}'}$ is the matrix of target output were
     each row represents the target output at the corresponding time $t$. 
Note that this also works for multi-dimensional output, in which case ${\bf W}^{out'}$
 will be a matrix containing connection weights between each pair of reservoir nodes and output nodes.

\begin{figure}[t]
\centering
\includegraphics[]{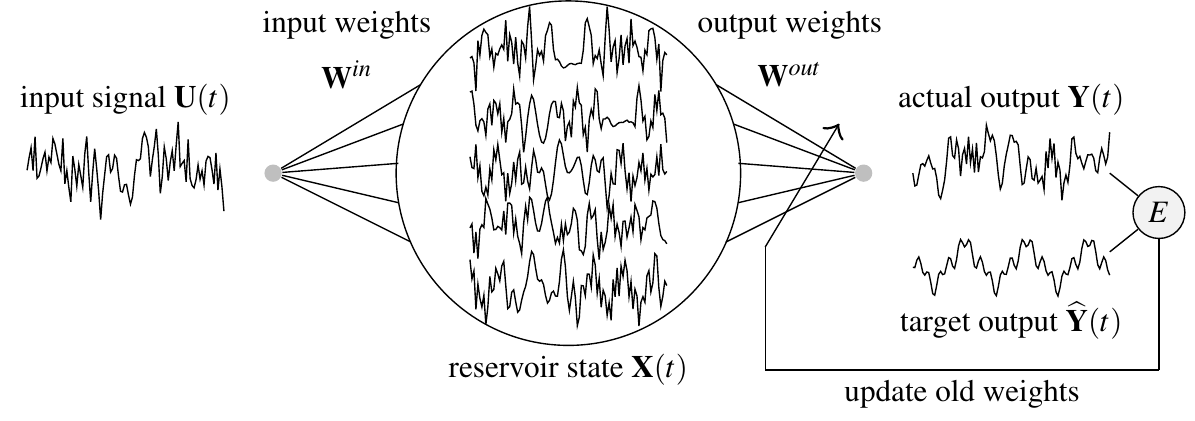}
\caption{Computation in a reservoir computer. The input signal ${\bf U}(t)$ is fed 
into every reservoir node $i$ with a corresponding weight $w^{in}_i$ 
denoted with weight column vector ${\bf W}^{in}=[w^{in}_i]$. Reservoir nodes are 
themselves coupled with each other using the weight matrix ${\bf W}^{res}
=[w^{res}_{ij}]$, where $w^{res}_{ij}$ is the weight of the connection from node $j$ to node $i$.}
\label{fig:fig0}
\end{figure}

Conceptually, the reservoir's role in RC is to act as a spatiotemporal kernel 
and project the input into a high-dimensional feature space 
\protect\cite{Lukosevicius:2009p1443}. In machine learning, this is usually referred 
to as feature extraction and is done to find  hidden structures 
in data sets or time series. The output is then calculated by properly weighting 
and combining different features of the data \protect\cite{bishop06prml}. An ideal 
reservoir should be able to perform feature extraction in a way that makes the 
mapping from feature space to output a linear problem. However, this is not 
always possible. In theory an ideal reservoir computer should have two 
features: a {\em separation property} of the reservoir and an {\em approximation 
property} of the readout layer. The former means the reservoir perturbations 
from two distinct inputs must remain distinguishable over time and the latter 
refers to the ability of the readout layer to map the reservoir state to a given 
target output in a sufficiently precise way.

Another way to understand computation in a high-dimensional recurrent systems is through analyzing their attractors. In this view, the state-space of the reservoir is partitioned into multiple basins of attraction. A basin of attraction is a subspace of the system's state-space, in which the system follows a trajectory towards its attractor. Thus computation takes place when the reservoir jumps between basins of attraction due to perturbations by an external input \protect\cite{Sussillo:2012fk,Sussillo:2009uq,PhysRevLett.108.128702,Krawitz:2007p1469}. On the other hand, one could directly analyze computation in the reservoir as the reservoir's average instantaneous information content to produce a desired output \protect\cite{snyder2013a}.

There has been much research to find 
the optimal reservoir structure and readout strategy. Jaeger 
\protect\cite{Jaeger:2001p1446} suggests that in addition to 
the separation property, the reservoir should have fading memory to forget past 
inputs after some period of time. He achieves this by adjusting the standard 
deviation of the reservoir weight matrix $\sigma$ so that the spectral radius 
of ${\bf W}^{res}$ remains close to 1, but slightly less than 1. This 
ensures that the reservoir can operate near critical dynamics, right at the 
edge between ordered and chaotic regimes. A key feature of critical systems is that 
perturbations to the system's trajectory neither spread nor die out, independent of the
system size\protect\cite{rohlf07_prl}, which makes adaptive information processing 
robust to noise\protect\cite{PhysRevLett.108.128702}. Other studies have also 
suggested that the critical dynamics is essential for good performance in RC 
\protect\cite{Natschlaeger2003,Bertschinger:2004p1450,snyder2013a,4905041020100501,Boedecker2009}.

 The RC architecture does not assume any specifics about the underlying 
reservoir. The only requirement is that it provides a suitable kernel 
to project inputs into a high-dimensional feature space. Reservoirs operating in the
 critical dynamical regime usually satisfy this requirement. Since 
RC makes no assumptions about the structure of the underlying 
reservoir, it is very suitable for use with unconventional computing 
paradigms \protect\cite{springerlink:10.1007,NIPS2012_0456,Paquot:2012uq}. Here,
 we propose and simulate a simple design for a reservoir computer based on a network of deoxyribozyme oscillators.

\section{Reservoir Computing using Deoxyribozyme Oscillators}
\label{sec:rcosci}
To make a DNA reservoir computer, we first need  a reservoir of DNA 
species with rich transient dynamics. To this end, we use a microfluidic 
reaction chamber in which different DNA species can interact. This 
must be an open reactor because we need to continually give input to the 
system and read its outputs. The reservoir state consists of the 
time-varying concentration of various species inside the chamber, and we compute
 using the reaction dynamics of the species inside the 
reactor. To perturb the reservoir we encode the time-varying input as fluctuations
 in the influx of species to the reactor. In 
\protect\cite{Morgan2005,Farfel2006},  a 
network of three deoxyribozyme NOT gates showed stable oscillatory 
dynamics in an open microfluidic reactor. We extend this work by designing a reservoir 
computer using deoxyribozyme-based oscillators and  investigating 
their information-processing capabilities. 

The oscillator dynamics in \protect\cite{Farfel2006} suffices as
an excitable reservoir. Ideally, the 
readout layer should also be implemented in a similar microfluidic setting 
and integrated with the reservoir. However, as a proof of concept we 
assume that we can read the reservoir state using fluorescent probes and 
calculate the output weights using software. 

The oscillator network in 
\protect\cite{Farfel2006} is described using a system of nine ordinary differential 
equations (ODEs), which simulate the details of a laboratory experiment of 
the proposed design. However, this model is mathematically unwieldy.
 We first reduce the oscillator ODEs in \protect\cite{Farfel2006} to a form  
 more amenable to mathematical analysis:

\begin{equation}
\begin{array}{llllllll}
\vspace{2mm}
\frac{d [P_1]}{d t} & = & h \beta [S_1] ([G_1]-[P_3])-\frac{e}{V}[P_1], & \text{     } & \frac{d [S_1]}{d t} & = & \frac{S^m_1}{V} -  h \beta [S_1] ([G_1]-[P_3])-\frac{e}{V}[S_1],\\
\vspace{2mm}
\frac{d [P_2]}{d t} & = & h \beta [S_2] ([G_2]-[P_1])-\frac{e}{V}[P_2], & & \frac{d [S_2]}{d t} & = & \frac{S^m_2}{V} - h \beta [S_2] ([G_2]-[P_1])-\frac{e}{V}[S_2],\\
\vspace{2mm}
\frac{d [P_3]}{d t} & = & h \beta [S_3]([G_3]-[P_2])-\frac{e}{V}[P_3], & & \frac{d [S_3]}{d t} & = & \frac{S^m_3}{V} -h \beta [S_3]([G_3]-[P_2])-\frac{e}{V}[S_3].
\end{array}
\label{eq:reduced3notwithinputsimplified}
\end{equation}

In this model, $[P_i]$, $[S_i]$, and $[G_i]$ are concentrations of 
three species of product molecules, three species of 
substrate molecules, and three species of 
gate molecules inside the reactor, and $S^m_i$ is the influx rate of 
$[S_i]$. The brackets $[$\,$\cdot$\,$]$ indicate chemical concentration and should not  be confused
 with the matrix notation introduced above. When explicitly talking about the concentrations 
 at time $t$, we use $P_i(t)$ and $S_i(t)$.
$V$ is the volume of the reactor, $h$ the fraction of the reactor chamber
 that is well-mixed, $e$ is the efflux rate, and $\protect\beta$ is the reaction rate 
 constant for the gate-substrate reaction, which is assumed to be identical for all gates 
 and substrates, for simplicity.

 To use this system as a reservoir we must ensure that it has transient or sustained oscillation.
  This can be easily analyzed by forming the Jacobian of the system.  Observing that all
  substrate concentrations reach an identical and constant value relative to their magnitude, we can
  focus on the dynamics of the product concentrations and write 
  an approximation to the Jacobian of the system as follows: 
  \protect\begin{equation}
 {\bf J}=
 \begin{bmatrix}
 \vspace{2mm}
 \frac{d[P_1]}{d[P_1]} &  \frac{d[P_1]}{d[P_2]} &  \frac{d[P_1]}{d[P_3]}\\
  \vspace{2mm}
  \frac{d[P_2]}{d[P_1]} &  \frac{d[P_2]}{d[P_2]}  &  \frac{d[P_2]}{d[P_3]}\\
   \frac{d[P_3]}{d[P_1]} &  \frac{d[P_3]}{d[P_2]} &  \frac{d[P_3]}{d[P_3]}
 \end{bmatrix}
 =
 \begin{bmatrix}
 \vspace{2mm}
  -\frac{e}{V} & -h\beta[S_1] & 0 \\
  \vspace{2mm}
   0 & -\frac{e}{V} & -h\beta[S_2] \\
   -h\beta[S_1]  & 0 & -\frac{e}{V}
  \end{bmatrix}
 \end{equation}
 Assuming that volume of the reactor $V$ and the reaction rate constant $\beta$ are given, the Jacobian 
 is  a function of only the efflux rate $e$ and the substrate concentrations $[S_i]$. The eigenvalues of the Jacobian
 are given by:
 \protect\begin{equation}
 \begin{array}{lll}
 \lambda_1 &=& -h\beta([S_1][S_2][S_3])^\frac{1}{3}-\frac{e}{V}\\
 \lambda_2 & =&  \frac{1}{2}h\beta([S_1][S_2][S_3])^\frac{1}{3}-\frac{e}{V} +\frac{\sqrt{3}}{2}h\beta([S_1][S_2][S_3])^\frac{1}{3}i \\
  \lambda_3 & =&  \frac{1}{2}h\beta([S_1][S_2][S_3])^\frac{1}{3}-\frac{e}{V} -\frac{\sqrt{3}}{2}h\beta([S_1][S_2][S_3])^\frac{1}{3}i 
 \end{array}
 \end{equation}
The existence of complex eigenvalues tells us that the system has oscillatory behavior near its critical points. The period of this oscillation is given by $T=2\pi\frac{\sqrt{3}}{2}h\beta([S_1][S_2][S_3])^{-\frac{1}{3}}$ and can be adjusted by setting appropriate base values for $S^m_i$. For sustained oscillation, the real part of the eigenvalues should be zero, which can be obtained by a combination of efflux rate and substrate influx rates such that $\frac{1}{2}h\beta([S_1][S_2][S_3])^\frac{1}{3}-\frac{e}{V}=0$.

\begin{figure}[t]
\centering
\subfloat[substrate in the reduced model]{
\includegraphics[width=2.2in]{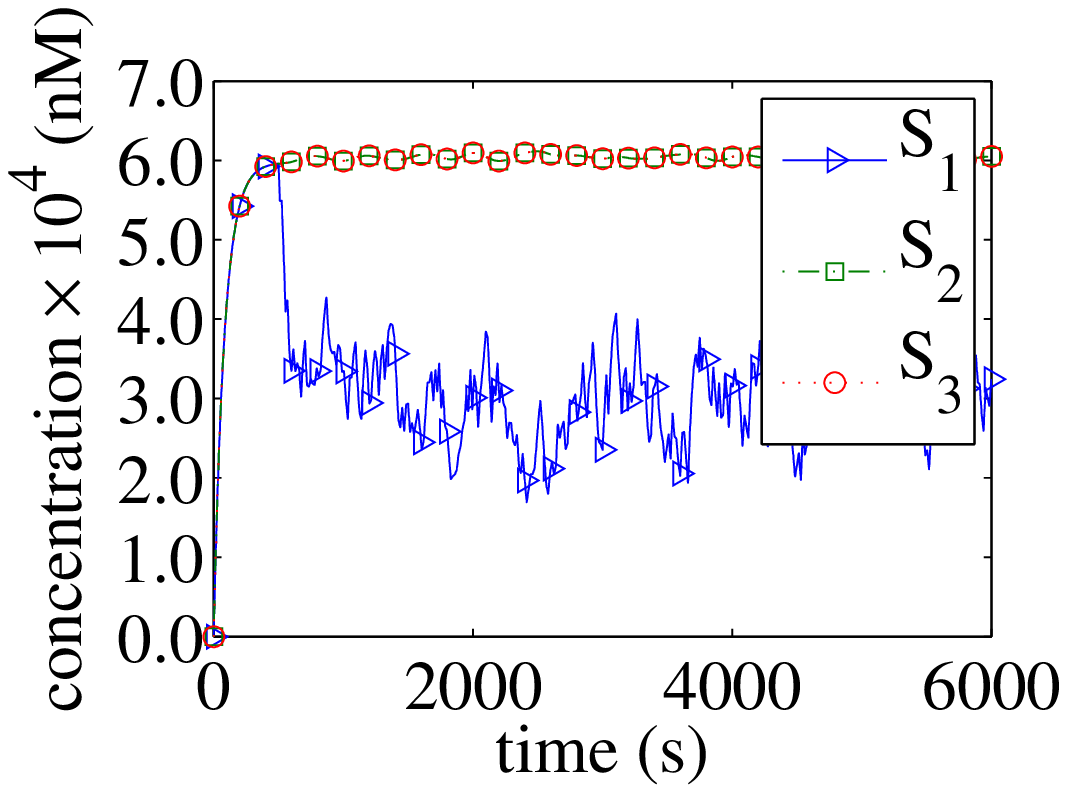}
\label{fig:subflucredo}
}\qquad
\subfloat[products in the reduced model]{
\includegraphics[width=2.2in]{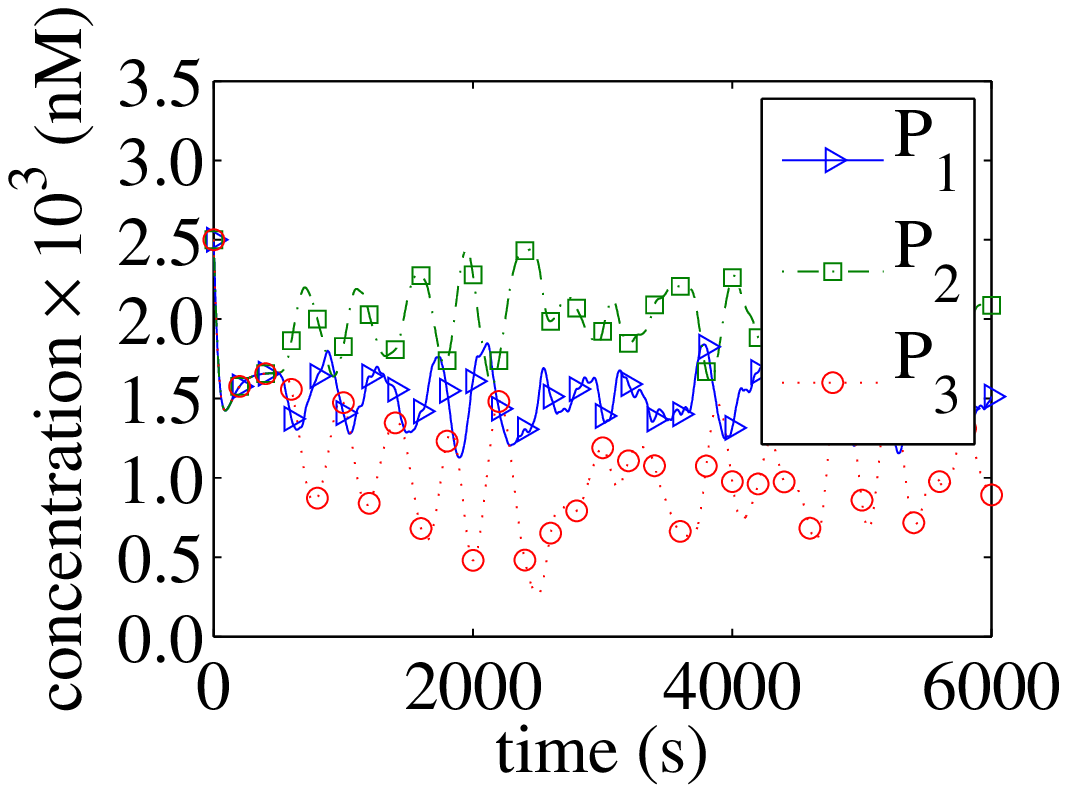}
\label{fig:prodflucredo}
}

\caption{The random fluctuation in substrate influx rate $S^m_1$ leaves 
traces on the oscillator 
dynamics that can be read off a readout layer. We observe the traces of the 
substate influx rate fluctuation both in the dynamics of the substrate
concentrations~\protect\subref{fig:subflucredo} and the product
concentrations~\protect\subref{fig:prodflucredo}. Both substrate and product
concentrations potentially carry information about the input. Substrate 
concentration $S_1$ is directly affected by $S^m_1$ and therefore shows very 
rapid fluctuations.}
\label{fig:fig4}
\end{figure}

\begin{figure}[b]
\centering
\includegraphics[]{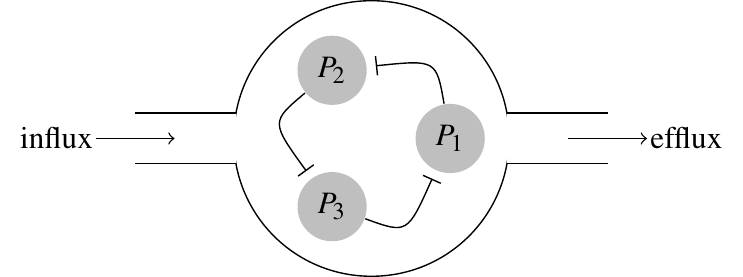}
\caption{Three products form an inhibitory cycle that  leads to oscillatory behavior 
in the reservoir. Each product $P_i$ inhibits the production of $P_{i+1}$ by the
 corresponding deoxyribozyme (cf. Equation~\ref{eq:reduced3notwithinputsimplified}).}
\label{fig:fig2}
\end{figure}

This model works as follows. The substrate molecules enter 
the reaction chamber and are bound to and cleaved by active gate molecules 
that are immobilized inside the reaction chamber, e.g., on beads. This reaction 
turns substrate molecules into the corresponding product molecule. However,
 the presence of each product molecule concentration suppresses the reaction
  of other substrates and gates. These three coupled reaction and inhibition cycles
   give rise to the oscillatory behavior of the products' concentrations
    (Figure~\protect\ref{fig:fig2}). Input is given to the system as fluctuation to one or more of
     the substrate influx rates. In Figure~\protect\ref{fig:subflucredo} we see that the
      concentration of $S_1$ varies rapidly as a response to the random fluctuations
       in $S^m_1$. This will result in antisymmetric 
concentrations of the substrate species inside the chamber and thus irregular 
oscillation of the concentration of product molecules. This irregular oscillation
 embeds features of the input fluctuation within it
 (Figure~\protect\ref{fig:subflucredo}). To keep the volume fixed, there is a continuous
   efflux of the chamber content. The Equation~\protect\ref{eq:reduced3notwithinputsimplified} assumes $([G_i]-[P_j])>0$, which should be taken into account while choosing initial concentrations and constants to simulate the system.

To perturb the intrinsic dynamics inside the reactor, an input signal can modulate 
one or more substrate influx rates. In our system, we achieve this by fluctuating
 $S^m_1$. In order to let the oscillators react to different values of $S^m_i$, 
we keep each new value of $S^m_i$ constant for $\tau$ seconds. In a basic 
setup, the initial concentrations of all the substrates inside the reactor are 
zero. Two of the product concentrations $P_2(0)$ and $P_3(0)$ are also 
set to zero, but to break the symmetry in the system and let the oscillation 
begin we set $P_1(0)=1000$ nM. The gate concentrations are set uniformly to $[G_i]=2500$ nM. This ensures that $([G_i]-[P_j])>0$ in our setup. The base values for substrate-influx rates are set 
to $5.45 \times 10^{-6}$ nmol\,s$^{-1}$.
Figure~\protect\ref{fig:fig4} shows the traces of computer simulation of this model, where $
\tau=30$ s. We use the reaction rate constant from \protect\cite{Farfel2006}, $\beta=5 \times 10^{-7}$ nM\,s$^{-1}$. Although the kinetics of immobilized deoxyribozyme may be different, for simplicity we use the reaction rate constant of free deoxyribozymes and we assume that we can find deoxyribozymes with appropriate kinetics when immobilized.  The values for the remaining constants are $e=8.8750 \protect\times 10^{-2}$ nL\protect\,s$^{-1}$ and $h=0.7849$, i.e., the average fraction of well-mixed solution calculated in \protect\cite{Farfel2006}. We assume the
 same microscale continuous stirred-tank reactor ($\mu$CSTR) as
  \protect\cite{Farfel2006,Morgan2005,Chou:2001fk}, which has volume $V=7.54$ nL. 
  The small volume of the reactor lets us achieve high concentration of oligonucleotides
   with small amounts of material; a suitable experimental setup is described
    in \protect\cite{C2LC40649G}.

The dynamics of the substrates (Figure~\ref{fig:subflucredo}) and products 
(Figure~\ref{fig:subflucredo}) are instructive as to what we can 
use as our reservoir state. Our focus will be the product concentrations. 
However, the substrate concentrations also show interesting irregular 
behavior that can potentially carry information about the input signals. This is 
not surprising since all of the substrate and product concentrations are 
connected in our oscillator network. However, the one substrate that is directly
 affected by the influx ($S_1$ in this case) shows the most intense fluctuations that 
 are directly correlated with the input. In some cases providing this extra information
  to the readout layer can help to find the right mapping between the reservoir state
   and the target output.

In the next section, we build two different reservoir computers using the 
dynamics of the concentrations in the reactor and use them to solve sample 
temporal tasks. Despite the simplicity of our system, it can 
retain the memory of past inputs inside the reservoir and use it to produce 
output. 

\section{Task  Solving Using a Deoxyribozyme Reservoir Computer}
\label{sec:tasksolving}

We saw in the preceding section that we can use substrate influx fluctuation as input 
to our molecular reservoir. We now show that we can train a 
readout layer to map the dynamics of the oscillator network to a target output. 
Recall that $\tau$ is the input hold time during which we 
keep $S^m_1$ constant so that the oscillators can react to different values of 
$S^m_1$. In other words, at the beginning of each $\tau$ interval a new 
random value for substrate influx is chosen and held fixed for $\tau$ seconds. 
Here, we set the input hold time $\tau=100$ s. In addition, 
before computing with the reservoir we must make sure that it has 
settled in its natural dynamics, otherwise the output layer will see dynamical 
behavior that is due to the initial conditions of the oscillators and not the input 
provided to the system. In the model, the oscillators reach their stable 
oscillation pattern within $500$ s. Therefore, we start our reservoir by 
using a fixed $S^m_1$ as described in Section~\ref{sec:rcosci} and run it for $500$ s before introducing fluctuations in $S^m_1$.

To study the performance of our DNA reservoir computer we use two different 
tasks, Task A and Task B, as toy problems. Both have recursive time dependence 
and therefore require the reservoir to remember past inputs for some 
period of time, and both are simplified versions of a popular RC 
benchmark, NARMA \protect\cite{Jaeger:2002p1445}. We define the 
input as $S^m_1(t)=S^{m*}_1R$, where $S^{m*}_1$ is the influx rate used 
for the normal working of the oscillators ($5.45\times 10^{-6}$ nmol\,s$^{-1}$ 
in our experiment) and $R$ is a random value between 0 and 1 sampled 
from a uniform distribution. We define the target output ${\bf \widehat{Y}(t)}$ of 
Task  A as follows:

\begin{equation}
{\bf \widehat{Y}}(t)=S^m_1(t-1)+2S^m_1(t-2).
\label{eq:taska}
\end{equation}

For  Task  B, we increase the length of the time dependence 
and make it a function of input hold time $\tau$. We define the target output 
as follows:

\begin{equation}
{\bf \widehat{Y}}(t)=S^m_1(t-\tau)+\frac{1}{2}S^m_1(t-\frac{3}{2}\tau).
\label{eq:taskb}
\end{equation}

\begin{figure}[t]
\centering
\includegraphics[width=4in]{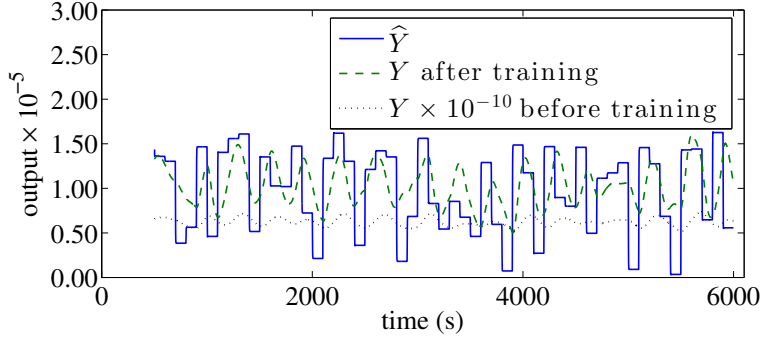}
\caption{Target output and the output of the molecular reservoir computer on  Task  A (Equation~\ref{eq:taska}) before and after training. After $500$ s the input starts to fluctuate randomly every $\tau$ seconds. In this example, the output of the system before training is 10 orders of magnitude larger than the target output. We rescaled the output before training to be able to show it in this plot. After training, the output is in the range of the target output and it tracks the fluctuations in the target output more closely.}
\label{fig:fig5}
\end{figure}

Note that the vectors ${\bf \widehat{Y}}(t)$ and ${\bf Y}(t)$ have only one row in this example. Figure~\protect\ref{fig:fig5} shows an example of the reservoir output ${\bf Y}(t)$ and the target output
 ${\bf \widehat{Y}}(t)$ calculated using Equation~\protect\ref{eq:taska} before and after training. In this example, the reservoir output before training is $10$ orders of magnitude off the target. 
 
 Our goal is to find a set of output weights so that ${\bf Y}(t)$ tracks the target output as closely as possible. We calculate the error using normalized root-mean-square error (NRMSE)  as follows:
 
 \begin{equation}
 \text{NRMSE} =\frac{1}{Y_{max}-Y_{min}} \sqrt{\frac{\sum_{t=t_1}^{t_n}({\bf \widehat{Y}}(t)-{\bf Y}(t))^2}{n}},
 \end{equation}
 where $Y_{max}$ and $Y_{min}$ are the maximum and the minimum of the ${\bf Y}(t)$ during the time interval $t_1<t<t_n$. The denominator $Y_{max}-Y_{min}$ is to ensure that $0\le \text{NRMSE} \le1$, where $\text{NRMSE}=0$ means ${\bf Y}(t)$ matches \protect${\bf \widehat{Y}}(t)$ perfectly.
 
Now we propose two different ways of calculating the output from the 
reservoir: (1) using only the dynamics of the product 
concentrations and (2) using both the product and  substrate concentrations. To formalize this using the block matrix notation, for the product-only version the reservoir state is given by ${\bf X}(t)={\bf P}(t)= [P_1(t) \text{ } P_2(t) \text{ }P_3(t)]^T$.  For the product-and-substrate version the reservoir state is given by vertically appending ${\bf S}(t)= [S_1(t) \text{ } S_2(t) \text{ } S_3(t)]^T$ to ${\bf P}(t)$, i.e., ${\bf X}(t)=[{\bf P}(t)\text{ } {\bf S}(t)]^T$, 
where ${\bf P}(t)$ is the column vector of the product concentrations as before 
and ${\bf S}(t)$ is the column vector of the substrate concentrations. We use $2000$ s of the reservoir dynamics ${\bf X}(t)$ to calculate the output weight matrix ${\bf W}^{out}$ using linear regression. We then test the generalization, i.e., how well the output ${\bf Y}(t)$ tracks the target ${\bf \widehat{Y}}(t)$ during  another $2000$ s period that we did not use to calculate the weights.

\begin{figure}[t]
\centering
\includegraphics[width=2in]{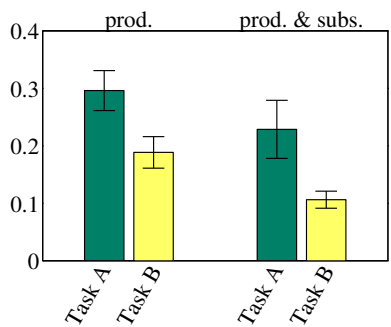}
\caption{Generalization NRMSE of the product-only and the product-and-substrate molecular reservoir computer on  
Task  A (Equation~\ref{eq:taska}) and Task B (Equation~\ref{eq:taskb}) averaged over $100$ trials. The bars and error bars show the mean and the standard deviation of NRMSE respectively.}
\label{fig:fig6}
\end{figure}

Figure~\protect\ref{fig:fig6} shows the mean and standard deviation of NRMSE of the reservoir computer using two different readout layer solving Task A and Task B. The product-and-substrate reservoir achieves a mean NRMSE of $0.23$ and $0.11$ on Task A and Task B with standard deviations $0.05$ and $0.02$ respectively, and the product-only reservoir achieves a mean NRMSE of $0.30$ and $0.19$ on Task A and Task B with standard deviations $0.04$ and $0.03$ respectively. As expected, the product-and-substrate reservoir computer achieves about $10\%$ improvement over the product-only version owing to its higher phase space dimensionality. Furthermore, both reservoirs achieve a $10\%$ improvement on Task B over Task A. This is surprising at first because Task B requires the reservoir to remember the input over a time interval of $\frac{3}{2} \tau$, but Task A only requires the last two time steps. However, to extract the features in the input signal, the input needs to percolate in the reservoir, which takes more than just two time steps. Task B requires more memory of the input, but also gives the reservoir enough time to process the input signal, which results in higher performance. Similar effects have been observed in~\protect\cite{snyder2013a}. Therefore, despite the very simple reservoir structure (three coupled oscillators), we can compute simple temporal tasks with $90\%$ accuracy. Increasing the number of oscillators and using the history of the oscillators dynamics similar to \cite{Appeltant:2011ys} could potentially lead to even higher performance.

\section{Discussion and Related Work}
\label{sec:discuss}

DNA chemistry is inherently programmable and highly 
versatile, and a number of different techniques have been developed, 
such as building digital and analog circuits using strand displacement 
cascades~\protect\cite{Qian03062011,Qian:2011uq}, developing 
game-playing molecular automata using deoxyribozymes~\protect
\cite{Pei:2010vn}, and directing self-assembly of nanostructures ~
\protect\cite{Yin:2008ys,Wei:2012ve,Ke30112012}. All of these 
approaches require precise design of DNA sequences to form the required 
structures and perform the desired computation. In this paper, we 
proposed a reservoir-computing approach to molecular computing. In nature, evidence for reservoir computing has been found in systems as simple as a bucket of water \protect\cite{Fernando2003}, simple organisms such as {\em E. Coli} \protect\cite{4218885}, and in systems as complex as the brain \protect\cite{Yamazaki2007290}. This approach does not require any specific 
behavior from the reactions, except that the reaction dynamics must 
result in a suitable transient behavior that we can use to compute
 \protect\cite{Lukosevicius:2009p1443}. This could give us a new perspective in long-term sensing, and potentially controlling, gene expression patterns over time in a cell. This would require appropriate sensors to detect cell state, for example the pH-sensitive DNA nanomachine recently reported by Modi et al. \protect\cite{Modi:2013ys}. This may result in new methods for smart diagnosis and treatment using DNA signal translators \cite{000205747600013n.d.,Beyer01012006,Shapiro17102008}.

 In RC, computation 
takes place as a transformation from the input space to a high-
dimensional spatiotemporal feature space created by the transient 
dynamics of the reservoir. Mathematical analysis suggests that all dynamical systems  show the 
same information processing capacity \protect\cite{Dambre:2012fk}. 
However, in practice, the performance of a reservoir is significantly 
affected by its dynamical regime. Many studies have shown that to 
achieve  a suitable reservoir in general, the underlying dynamical 
system must operate in the critical dynamical regime \protect
\cite{snyder2013a,Jaeger:2002p1445,Bertschinger:
2004p1450,4905041020100501}.

We used the dynamics of the concentrations of different 
molecular species to extract features of an input signal and map them 
to a desired output. As a proof of concept, we proposed a reservoir 
computer using deoxyribozyme oscillator network and showed how to 
provide it with input and read its outputs. However, in our setup, we 
assumed that we read the reservoir state using fluorescent probes 
and process them using software. In principle, the mapping from the 
reservoir state to target output can be carried out as an integrated 
part of the chemistry using an approach similar to the one reported in
\protect\cite{Qian:2011uq}, which implements a neural network using 
strand displacement. In\protect\cite{LakinMR:towblm}, we  
proposed a  chemical reaction network  inspired by deoxyribozyme 
chemistry  that can learn a linear function  and repeatedly use it to 
classify input signals. In principle, these methods could be used to 
implement the regression algorithm and therefore the readout layer 
as an integrated part of the molecular reservoir computer. A microfluidic reactor has been demonstrated in~\protect\cite{C2LC40649G} that would be suitable for implementing our system. Therefore, the molecular reservoir computer that we proposed here is physically plausible and can be  implemented in the laboratory using microfuidics.

\section{Conclusion and Future Work}
\label{sec:conc}
We have proposed and simulated a novel approach to DNA computing 
based on the  reservoir computing paradigm. Using 
a network of oscillators built from deoxyribozymes we can extract 
hidden features in a given input signal and compute any desired 
output. We tested the performance of this approach on two simple 
temporal tasks. This approach is generalizable to different molecular 
species so long as  they possess rich reaction 
dynamics. Given the available 
technology today this approach is plausible and can lead to many 
innovations in biological signal processing, which has important applications in smart 
diagnosis and treatment techniques.
In future work, we shall study the use of other sets of reactions for the reservoir. Moreover, for any real-world application of this technique, we have to address  the chemical implementation of the readout layer. An important open question is the complexity of molecular reactions necessary to achieve critical dynamics in the reservoir. For practical applications, the effect of sparse input and sparse readout needs thorough investigation, i.e., how should one distribute the input to the reservoir and how much of the reservoir dynamics is needed for the readout layer to reconstruct the target output accurately? It is also possible to use the history of the reservoir dynamics to compute the output, which would require addition of a feedback channel to the reactor. The molecular readout layer could be set up to read the species concentration along the feedback channel. Another possibility is to connect many reactors to create a modular molecular reservoir computer, which could be used strategically to scale up to more complex problems.

%
%

\bibliographystyle{splncs}
\bibliography{dna19}

\begin{thebibliography}{10}

\bibitem{Jaeger02042004}
Jaeger, H., Haas, H.:
\newblock Harnessing nonlinearity: Predicting chaotic systems and saving energy
  in wireless communication.
\newblock Science \textbf{304}(5667) (2004)  78--80

\bibitem{Farfel2006}
Farfel, J., Stefanovic, D.:
\newblock Towards practical biomolecular computers using microfluidic
  deoxyribozyme logic gate networks.
\newblock In Carbone, A., Pierce, N., eds.: DNA Computing. Volume 3892 of
  Lecture Notes in Computer Science.
\newblock Springer Berlin Heidelberg (2006)  38--54

\bibitem{springerlink:10.1007}
Luko{\v s}evi{\v c}ius, M., Jaeger, H., Schrauwen, B.:
\newblock Reservoir computing trends.
\newblock KI - K{\"u}nstliche Intelligenz \textbf{26}(4) (2012)  365--371

\bibitem{NIPS2012_0456}
Smerieri, A., Duport, F., Paquot, Y., Schrauwen, B., Haelterman, M., Massar,
  S.:
\newblock Analog readout for optical reservoir computers.
\newblock In Bartlett, P., Pereira, F., Burges, C., Bottou, L., Weinberger, K.,
  eds.: Advances in Neural Information Processing Systems 25: 26th Annual
  Conference on Neural Information Processing Systems.
\newblock Curran Associates, Inc. (2012)  953--961

\bibitem{Paquot:2012uq}
Paquot, Y., Duport, F., Smerieri, A., Dambre, J., Schrauwen, B., Haelterman,
  M., Massar, S.:
\newblock Optoelectronic reservoir computing.
\newblock Scientific Reports \textbf{2} (2012)

\bibitem{Lukosevicius:2009p1443}
Luko{\v s}evi{\v c}ius, M., Jaeger, H.:
\newblock Reservoir computing approaches to recurrent neural network training.
\newblock Computer Science Review \textbf{3}(3) (2009)  127--149

\bibitem{Maass:2002p1444}
Maass, W., Natschl{\"a}ger, T., Markram, H.:
\newblock Real-time computing without stable states: a new framework for neural
  computation based on perturbations.
\newblock Neural computation \textbf{14}(11) (2002)  2531--60

\bibitem{Jaeger:2002p1445}
Jaeger, H.:
\newblock {Tutorial on training recurrent neural networks, covering BPPT, RTRL,
  EKF and the ‚``echo state network‚" approach}.
\newblock Technical Report GMD Report 159, German National Research Center for
  Information Technology, St. Augustin-Germany (2002)

\bibitem{Widrow:1990p2053}
Widrow, B., Lehr, M.:
\newblock 30 years of adaptive neural networks: Perceptron, madaline, and
  backpropagation.
\newblock Proceedings of the IEEE \textbf{78}(9) (1990)  1415--1442

\bibitem{PSP:2043984}
Penrose, R.:
\newblock A generalized inverse for matrices.
\newblock Mathematical Proceedings of the Cambridge Philosophical Society
  \textbf{51} (1955)  406--413

\bibitem{bishop06prml}
Bishop, C.M.:
\newblock Pattern Recognition and Machine Learning (Information Science and
  Statistics).
\newblock Springer-Verlag New York, Inc., Secaucus, NJ, USA (2006)

\bibitem{Sussillo:2012fk}
Sussillo, D., Barak, O.:
\newblock Opening the black box: Low-dimensional dynamics in high-dimensional
  recurrent neural networks.
\newblock Neural Computation \textbf{25}(3) (2012)  626--649

\bibitem{Sussillo:2009uq}
Sussillo, D., Abbott, L.F.:
\newblock Generating coherent patterns of activity from chaotic neural
  networks.
\newblock Neuron \textbf{63}(4) (2009)  544--557

\bibitem{PhysRevLett.108.128702}
Goudarzi, A., Teuscher, C., Gulbahce, N., Rohlf, T.:
\newblock Emergent criticality through adaptive information processing in
  {B}oolean networks.
\newblock Phys. Rev. Lett. \textbf{108} (2012)  128702

\bibitem{Krawitz:2007p1469}
Krawitz, P., Shmulevich, I.:
\newblock Basin entropy in boolean network ensembles.
\newblock Phys. Rev. Lett. \textbf{98}(15) (2007)  158701

\bibitem{snyder2013a}
Snyder, D., Goudarzi, A., Teuscher, C.:
\newblock Computational capabilities of random automata networks for reservoir
  computing.
\newblock Phys. Rev. E \textbf{87} (2013)  042808

\bibitem{Jaeger:2001p1446}
Jaeger, H.:
\newblock Short term memory in echo state networks.
\newblock Technical Report GMD Report 152, GMD-Forschungszentrum
  Informationstechnik (2002)

\bibitem{rohlf07_prl}
Rohlf, T., Gulbahce, N., Teuscher, C.:
\newblock Damage spreading and criticality in finite random dynamical networks.
\newblock Phys. Rev. Lett. \textbf{99}(24) (2007)  248701

\bibitem{Natschlaeger2003}
Natschl{\"a}ger, T., Maass, W.:
\newblock Information dynamics and emergent computation in recurrent circuits
  of spiking neurons.
\newblock In Thrun, S., Saul, L., Schoelkpf, B., eds.: Proc. of NIPS 2003,
  Advances in Neural Information Processing Systems. Volume~16., Cambridge, MIT
  Press (2004)  1255--1262

\bibitem{Bertschinger:2004p1450}
Bertschinger, N., Natschl{\"a}ger, T.:
\newblock Real-time computation at the edge of chaos in recurrent neural
  networks.
\newblock Neural Computation \textbf{16}(7) (2004)  1413--1436

\bibitem{4905041020100501}
B{\"u}sing, L., Schrauwen, B., Legenstein, R.:
\newblock Connectivity, dynamics, and memory in reservoir computing with binary
  and analog neurons.
\newblock Neural Computation \textbf{22}(5) (2010)  1272--1311

\bibitem{Boedecker2009}
Boedecker, J., Obst, O., Mayer, N.M., Asada, M.:
\newblock Initialization and self-organized optimization of recurrent neural
  network connectivity.
\newblock HFSP Journal \textbf{3}(5) (2009)  340--349

\bibitem{Morgan2005}
Morgan, C., Stefanovic, D., Moore, C., Stojanovic, M.N.:
\newblock Building the components for a biomolecular computer.
\newblock In Ferretti, C., Mauri, G., Zandron, C., eds.: DNA Computing. Volume
  3384 of Lecture Notes in Computer Science.
\newblock Springer Berlin Heidelberg (2005)  247--257

\bibitem{Chou:2001fk}
Chou, H.P., Unger, M., Quake, S.:
\newblock A microfabricated rotary pump.
\newblock Biomedical Microdevices \textbf{3}(4) (2001)  323--330

\bibitem{C2LC40649G}
Galas, J.C., Haghiri-Gosnet, A.M., Estevez-Torres, A.:
\newblock A nanoliter-scale open chemical reactor.
\newblock Lab Chip \textbf{13} (2013)  415--423

\bibitem{Appeltant:2011ys}
Appeltant, L., Soriano, M.C., Van~der Sande, G., Danckaert, J., Massar, S.,
  Dambre, J., Schrauwen, B., Mirasso, C.R., Fischer, I.:
\newblock Information processing using a single dynamical node as complex
  system.
\newblock Nature Communications \textbf{2} (2011)

\bibitem{Qian03062011}
Qian, L., Winfree, E.:
\newblock Scaling up digital circuit computation with {DNA} strand displacement
  cascades.
\newblock Science \textbf{332}(6034) (2011)  1196--1201

\bibitem{Qian:2011uq}
Qian, L., Winfree, E., Bruck, J.:
\newblock Neural network computation with {DNA} strand displacement cascades.
\newblock Nature \textbf{475}(7356) (2011)  368--372

\bibitem{Pei:2010vn}
Pei, R., Matamoros, E., Liu, M., Stefanovic, D., Stojanovic, M.N.:
\newblock Training a molecular automaton to play a game.
\newblock Nature Nanotechnology \textbf{5}(11) (2010)  773--777

\bibitem{Yin:2008ys}
Yin, P., Choi, H.M.T., Calvert, C.R., Pierce, N.A.:
\newblock Programming biomolecular self-assembly pathways.
\newblock Nature \textbf{451}(7176) (2008)  318--322

\bibitem{Wei:2012ve}
Wei, B., Dai, M., Yin, P.:
\newblock Complex shapes self-assembled from single-stranded {DNA} tiles.
\newblock Nature \textbf{485}(7400) (2012)  623--626

\bibitem{Ke30112012}
Ke, Y., Ong, L.L., Shih, W.M., Yin, P.:
\newblock Three-dimensional structures self-assembled from {DNA} bricks.
\newblock Science \textbf{338}(6111) (2012)  1177--1183

\bibitem{Fernando2003}
Fernando, C., Sojakka, S.:
\newblock Pattern recognition in a bucket.
\newblock In Banzhaf, W., Ziegler, J., Christaller, T., Dittrich, P., Kim, J.,
  eds.: Advances in Artificial Life. Volume 2801 of Lecture Notes in Computer
  Science.
\newblock Springer Berlin Heidelberg (2003)  588--597

\bibitem{4218885}
Jones, B., Stekel, D., Rowe, J., Fernando, C.:
\newblock Is there a liquid state machine in the bacterium \textit{Escherichia
  coli}?
\newblock In: Artificial Life, 2007. ALIFE '07. IEEE Symposium on. (2007)
  187--191

\bibitem{Yamazaki2007290}
Yamazaki, T., Tanaka, S.:
\newblock The cerebellum as a liquid state machine.
\newblock Neural Networks \textbf{20}(3) (2007)  290--297

\bibitem{Modi:2013ys}
Modi, S., Nizak, C., Surana, S., Halder, S., Krishnan, Y.:
\newblock Two {DNA} nanomachines map {pH} changes along intersecting endocytic
  pathways inside the same cell.
\newblock Nat Nano \textbf{8}(6) (2013)  459--467

\bibitem{000205747600013n.d.}
Beyer, S., Dittmer, W., Simmel, F.:
\newblock Design variations for an aptamer-based {DNA} nanodevice.
\newblock Journal of Biomedical Nanotechnology \textbf{1}(1) (2005)  96--101

\bibitem{Beyer01012006}
Beyer, S., Simmel, F.C.:
\newblock A modular {DNA} signal translator for the controlled release of a
  protein by an aptamer.
\newblock Nucleic Acids Research \textbf{34}(5) (2006)  1581--1587

\bibitem{Shapiro17102008}
Shapiro, E., Gil, B.:
\newblock {RNA} computing in a living cell.
\newblock Science \textbf{322}(5900) (2008)  387--388

\bibitem{Dambre:2012fk}
Dambre, J., Verstraeten, D., Schrauwen, B., Massar, S.:
\newblock Information processing capacity of dynamical systems.
\newblock Scientific Reports \textbf{2} (2012)

\bibitem{LakinMR:towblm}
Lakin, M.R., Minnich, A., Lane, T., Stefanovic, D.:
\newblock {T}owards a biomolecular learning machine.
\newblock In Durand-Lose, J., Jonoska, N., eds.: {U}nconventional {C}omputation
  and {N}atural {C}omputation 2012. Volume 7445 of Lecture Notes in Computer
  Science., Springer-Verlag (2012)  152--163

\end{thebibliography}

\end{document}